# Counting Solutions of Constraint Satisfiability Problems: Exact Phase Transitions and Approximate Algorithm

**Minghao Yin, Huang Ping**

**Department of Computer, Northeast Normal University, Changchun, China, 130117**

**ymh@nenu.edu.cn**

**Abstract**

The study of phase transition phenomenon of NP complete problems plays an important role in understanding the nature of hard problems. In this paper, we follow this line of research by considering the problem of counting solutions of Constraint Satisfaction Problems (#CSP). We consider the random model, i.e. RB model. We prove that phase transition of #CSP does exist as the number of variables approaches infinity and the critical values where phase transitions occur are precisely located. Preliminary experimental results also show that the critical point coincides with the theoretical derivation. Moreover, we propose an approximate algorithm to estimate the expectation value of the solutions number of a given CSP instance of RB model.

## Introduction

In the past decade there has been a significant interest in the phase transition of NP-complete or NP-hard problems. Phase transition is usually a transformation from one state to another state suddenly when a particular parameter is varied. For instance, propositional satisfiablity problems (SAT), the parameter controlled for phase transition is the density of clause (i.e. number of clauses/number of variables). However, it seems that it is always difficult to obtain the location of the exact phase transition point. Still take SAT for the example, from the experimental aspect, (Mitchell et al. 1992) pointed out that the phase transition point is approximately 4.3 by analyzing the experimental results. (Kirkpatrick and Selman 1994) showed that the phase transition point should be 4.17 by using a statistical physics method. From the theoretical aspect, the obtained phase transition points are usually in the form of some loose but hard won bounds. Until now, within our knowledge, the best lower bound and upper bound of SAT are 3.52 (Kaporis el al. 2006) and 4.49 (Diaz et al. 2008) respectively. For constraint satisfaction problems (CSPs), a generalization of SAT, the locations of phase transition points have also been widely studied both from an experimental and a theoretical view. Among them, (Xu and Li 2000) may be one of the few works that proved the existence of phase transition and identified the phase transition points exactly. In that literature, a new type of random

CSP model, i.e. Model RB, is introduced. Model RB is a revision of the standard random Model B and has some nice features. First, the existence of phase transition can be proved and the exact location of phase transition points can be quantified. Second, it is shown that both theoretically and experimentally Model RB can be used to generate hard satisfiable instances by translating CSPs into CNF formulas (Xu el al. 2007).

In fact, the phase transition phenomenon does not only occur in NP-complete problems or NP-hard, but also occurs in problems of complex classes beyond NP. For instance, researchers have already investigated the phase transition in QSAT and classical planning problems (Bylander 1996, Gent and Walsh 1999). On the other hand, in recent years we have also viewed an increasing interest in counting the number of solutions to combinatorial problems. Recent research has shown that counting the model of SAT problems or CSP problems corresponds to numerous #P-complete problems such as conformant probabilistic planning (Domshlak and Hoffman 2006), and various probabilistic inference problems (Chavira and Darwiche 2008). (Bailey et al. 2007) first studied phase transition in model counting problems of SAT. They considered the decision problem #3SAT($\geq 2^{n/2}$): given a 3CNF-formula, is it satisfied by at least the square-root of the total number of possible truth assigment. From experimental results, Bailey et al. (2007) pointed out that the phase transition point should be 2.5. And from theoretical aspect, they proved that the lower bound and the upper bound should be 0.9227 and 2.595.

The first aim of this paper is to extend this line of research to study phase transition of a generalization of #SAT, i.e. counting the solutions of constraint satisfaction problems. Similar to (Bailey et al. 2007) considering a decision version of #SAT, we consider a decision version of #CSP, as called #CSP ($\geq d^{n/t}$). Specifically, for every integer $t \geq 2$, given a CSP instance with $n$ variables and let the cardinality of the domain set be $d$, #CSP ($\geq d^{n/t}$) is the problem of deciding whether the instance has at least $d^{n/t}$ satisfying assignments. Particularly, for t=2, we have the decision problem #CSP ($\geq d^{n/2}$): given a CSP instance, is it satisfied by at least the square-root of the total number of possible assignments? Note that #CSP ($\geq d^{n/t}$) can be viewed as a generalization of #3SAT($\geq 2^{n/t}$), therefore #CSP ($\geq d^{n/t}$) is at least PP-hard, and so is #CSP ($\geq d^{n/2}$). We prove that due to incorporating uniformly variant domain size, phase transition does exist for Model RB in #CSP ($\geq d^{n/t}$) problems. Moreover, the critical value of the phase transition point can be also quantified for #CSP ($\geq d^{n/t}$) problems.

Another active research area in counting solution number of combinatorial problems is to develop efficient algorithms for these problems. For #SAT, (Birnbaum and Lozinskii 1999) first proposed an exact algorithm CDP, an extension of the standard Davis-Putnam(DP) procedure, to solve #SAT problems. After that, different kinds of skills like component caching and clause learning are incorporated to enhance the counting ability of #SAT (Gomes et. al. 2009). Since the run time and memory usage of exact algorithms usually increase exponentially with problem size, approximate algorithm are introduced, such as Approx-count, Sample-count and Sample-Minisat algorithms. For #CSP, Gomes et. al. (2007) considered binary and generalized XOR constraints, and developed efficient complete domain filtering algorithms. In this paper, we propose an approximate algorithm for #CSP. Interestingly, first, this algorithm only needs linear time, which means that the computing process is very efficient. Second, unlike other approximate algorithms, the algorithm becomes more accurate when the problem scale increases. We can prove that this method can obtain good bounds of the solution numbers of Model RB with high confidence.

## Preliminary

In this paper, a *constraint satisfaction problem* (CSP) $P$ can be viewed as a triple $(X, D, C)$ where $X=(x_1, x_2, ..., x_n)$ is a set of n variables. $D$ is a mapping from $X$ to a set of *domains* $D = (D(x_1), D(x_2), ..., D(x_n))$, where $D(x_i)$, $D_i$ for short, is the finite domain of its possible values. For $2 \leq k \leq n$ a *constraint* $c_{i_1, i_2, ..., i_k} \in C$ is defined as a pair $(X_i, R_i)$ such that $X_i = (x_{i_1}, ..., x_{i_k})$ is a subset of X called the *constraint scope* and $R_i$, as called *constraint relations*, is a subset of the Cartesian Product $D_{i_1} \times ... \times D_{i_k}$ and specifies allowed combinations of values for the variables in $X_i$. Given a CSP instance P, if there is a map from $X$ to the disjoint Union $\cup_{i=1}^{n} D_i$ with each $f(x_i) \in D_i$ satisfying that $f(X_i) = (f(x_{i_1}), ..., f(x_{i_k})) \in R_i$ for all $i=1,...,t$, then $P$ is satisfiable and we would say $f = (f(x_1), f(x_2), ..., f(x_n))$ is a *solution* of the instance P; otherwise, *P is unsatisfiable*. Each random CSP instance can be also described by a tuple $(k, n, d, p_1, p_2)$ where $k$ denotes the arity of the constraints, $n$ denotes the number of variables, $d$ denotes the uniform domain size, $p_1$ denotes the density of constraint and $p_2$ denotes the tightness of the constraints.

**Definition 1** (Model B) *A class of random CSP instances of model B will be denoted as a tuple* $(k, n, d, p_1, p_2)$ *where, for each instance:*

*1) $k \geq 2$ denotes the arity of each constraint,*

*2) $n \geq 2$ denotes the number of variables,*

*3) $d \geq 2$ denotes the size of each domain,*

*4) $1 \geq p_1 > 0$ determines the number of $m = p_1 \binom{n}{k}$ constraints,*

*5) $1 > p_2 > 0$ determines the number $t = p_2 d^k$ of disallowed tuples of each relation.*

(Achlioptas et. al. 2000) pointed out that the instances generated using Model B suffer from (trivial) insolubility when problem size increases. And therefore Xu and Li (2000) introduced an alternative random model, i.e. Mode RB.

**Definition 2** (*Model RB*) *A class of random CSP instances of model RB will be denoted as a tuple (k, n, α, r, p) where, for each instance:*

*1) k≥2 denotes the arity of each constraint,*

*2) n≥2 denotes the number of variables,*

*3) α>0 determines the domain size $d=n^\alpha$ of each domain,*

*4) r>0 determines the number $m = r \cdot n \cdot \ln n$ of constraints,*

*5) 1>p>0 determines the number $t = pd^k$ of disallowed tuples of each relation.*

The main difference between Model RB and Model B is that the domain size in Model RB grows with the number of variables. The generation of random CSP instance in Model RB is done as follows:

(1) Select $m = r \cdot n \cdot \ln n$ random constraints (with repetition), each one formed by randomly selecting k of n variables (without repetition).

(2) For each constraint select $t = pd^k$ incompatible tuples of values (without repetition). i.e., each constraint relation contains exactly $(1-p)d^k$ compatible tuples of values.

In order to determine the critical value where the phase transition occurs, we need the following definitions:

**Definition 3** (*Assignment pair*, Xu and Li 2000) *An assignment pair is an ordered pair <a, b> of assignments to the variables in U, where $a = (a_1, a_2, ..., a_n)$ and $b = (b_1, b_2, ..., b_n)$ with $a_l, b_l \in D_l$. An assignment pair <a, b> satisfies a CSP if and only if both a and b satisfies this CSP. The set that consist of all the assignment pairs is denoted by $A_{pair}$.*

**Definition 4** (*Similarity number*)

$S: A_{pair} \mapsto \{0,1,2,...\}$,

$$S(<a,b>) = n - ham(a, b)$$

*Where the function ham(a, b) is the Hamming distance between a and b,* defined as

$$ham(a,b) = \sum_{i=1}^{n} |a_i - b_i|$$

**Definition 5** (*Similarity degree*) s: $A_{pair} \mapsto R$,

$$s(<a, b>) = \frac{S(<a, b>)}{n}.$$

The similarity degree of an assignment pair is a measure how an assignments in the assignment pair is similar to the other. From the definition, we can see that $0 \leq s(<a, b>) \leq 1$, and the larger the similarity degree is, the more similar are the two assignments.

## Phase transitions of #CSP($\geq d^{n/t}$)

Xu and Li (2000) have proved that for CSP problems, an asymptotic phase transition can be guaranteed in Model RB with a limited restriction on domain size and on constraint tightness. In this section, we further prove that for #CSP($\geq d^{n/t}$), the asymptotic phase transition also exists, and the threshold can be precisely located as well. Let $Pr$ be the probabilistic distribution and let $X_{r,p}^{n,k,\alpha}$ denote the solutions number of the instance generated following Model RB. In this section, the following theorems are proved.

**Theorem 1** If $p_{cr}=1-e^{-\frac{\alpha}{r}(1-\frac{1}{t})}$, where $\alpha > 1/k$, $r>0$ are two constraints, $k$, $\alpha$ and $r$ satisfy the inequality $k \cdot e^{-\alpha/r} \geq 1$, then

$$\lim_{n \to \infty} \Pr[X_{r,p}^{n,k,\alpha} \geq d^{n/t}] = 1, \text{when } p < p_{cr} \tag{1}$$

$$\lim_{n \to \infty} \Pr[X_{r,p}^{n,k,\alpha} \geq d^{n/t}] = 0, \text{when } p > p_{cr} \tag{2}$$

*Proof.* First, according to the definition of Model RB, given an instance $I$ of Model RB, the expected number of solutions $E(X_{r,p}^{n,k,\alpha})$ of the instance is as follows:

$$E(X_{r,p}^{n,k,\alpha}) = d^n (1-p)^{rn \ln n} = n^{\alpha n}(1-p)^{rn \ln n}. \tag{3}$$

Then according to Markov's inequality,

$$\Pr[X_{r,p}^{n,k,\alpha} \geq 1] \leq E(X_{r,p}^{n,k,\alpha})$$

let $\theta$ be an arbitrary real number

$$\Pr[X_{r,p}^{n,k,\alpha}/\theta \geq 1] \leq E(X_{r,p}^{n,k,\alpha}/\theta).$$

Rewriting the formula, we get

$$\Pr[X_{r,p}^{n,k,\alpha} \geq \theta] \leq E(X_{r,p}^{n,k,\alpha})/\theta \tag{4}$$

Then according to formula (3) and (4), we can easily prove that $\lim_{n \to \infty} \Pr[X_{r,p}^{n,k,\alpha} \geq d^{n/t}] = 0$ holds when $p > p_{cr}$. So we finish the proof of equation (2).

The proof of equation (1) is more complicate. Let $a=(a_1, a_2, ..., a_n)$ and $b = (b_1, b_2, ..., b_n)$ stand for two assignments to variables of $I$. The key point of the proof is to derive $E(X_{r,p}^{n,k,\alpha} | a=1)$ and give an asymptotic estimate of it. Here $E(X_{r,p}^{n,k,\alpha} | a=1)$ denotes the conditional expectation of solutions number of the instance $I$ when $I$ is satisfied by $a$. Let $Pr(<a=1,b=1>)$ stand for the probability of an assignment pair $<a, b>$ satisfying $I$. Since each constraint is generated independently, to compute $Pr(<a=1,b=1>)$, we only need to consider the probability of the assignment pair $<a, b>$ satisfying a random constraint. Let the similarity number of $<a, b>$ be $S$, then:

(a) If assignment a is the same as b, the probability of $<a, b>$ satisfying the constraint is

$$\binom{d^k-1}{(1-p)d^K-1} \Big/ \binom{d^k}{(1-p)d^K} = 1-p.$$

The probability that a random constraint falls into this case is

$$\binom{S}{k} \Big/ \binom{n}{k}.$$

(b) Otherwise, the probability of $<a, b>$ satisfying the constraint is

$$\binom{d^k-2}{(1-p)d^K-2} \Big/ \binom{d^k}{(1-p)d^K} = (1-p)^2 + O(\frac{1}{d^k}).$$

The probability a random constraint falls into this case is

$$1-\binom{S}{k} \Big/ \binom{n}{k}.$$

From above two situations (a) and (b), we can draw the conclusion that

$Pr(<a=1,b=1>)$

$$= \left( (1-p)\cdot\binom{S}{k}\Big/\binom{n}{k} + (1-p)^2\cdot(1-\binom{S}{k}\Big/\binom{n}{k}) + O(\frac{1}{d^k}) \right)^m \tag{5}$$

Since all the constraints are selected independently, and according to the conditional probability, we have

$Pr(<a=1|b=1>) = Pr(<b=1|a=1>)$

$$= \left( \binom{S}{k}\Big/\binom{n}{k} + (1-p)\cdot(1-\binom{S}{k}\Big/\binom{n}{k}) + O(\frac{1}{d^k}) \right)^m \tag{6}$$

Accordingly, we have:
$$E(X_{r,p}^{n,k,\alpha}|a=1) = \sum_i Pr(i=1|a=1)$$
$$= \sum_{S=0}^{n} \binom{n}{S}(d-1)^{n-S}\left( \binom{S}{k}\Big/\binom{n}{k} + (1-p)(1-\binom{S}{k}\Big/\binom{n}{k}) + O(\frac{1}{d^k}) \right)^m$$

Here *i* denotes an arbitrary assignment of the instance I. It is easy to prove that the following inequality stands:

$$\Pr[X_{r,p}^{n,k,\alpha} \geq \theta] \geq \sum_i \frac{\pr(i=1)}{E(X_{r,p}^{n,k,\alpha}|i=1)} \tag{7}$$

Simplifying the formula, we have

$$\Pr[X_{r,p}^{n,k,\alpha} \geq \theta] \geq E(X_{r,p}^{n,k,\alpha}) / E(X_{r,p}^{n,k,\alpha}|a=1) \tag{8}$$

Note that $d = n^\alpha$ and $\alpha > 1/k$, we have

$$\binom{S}{k} / \binom{n}{k} = \binom{ns}{k} / \binom{n}{k} = s^k + g(s)/n + O(1/n^2) \tag{9}$$

According to Formula(3) and (8),

$$\Pr[X_{r,p}^{n,k,\alpha} \geq d^{n/t}]$$

$$\geq \frac{d^n(1-p)^m}{\sum_{S=0}^{n} \binom{n}{S}(d-1)^{n-S}((\binom{S}{k}/\binom{n}{k}) + (1-p).(1 - \binom{S}{k}/\binom{n}{k}) + O(\frac{1}{d^k}))^m} \tag{10}$$

For simplicity, denote the main part of (10) by *F(s)*

$$F(s) = \binom{n}{ns}(d-1)^{n-ns}((\binom{ns}{k}/\binom{n}{k}) + (1-p).(1 - \binom{ns}{k}/\binom{n}{k}) + O(\frac{1}{d^k}))^m$$

$$= \binom{n}{ns}(d-1)^{n-S}(s^k + \frac{g(s)}{n} + (1-p).(1 - s^k - \frac{g(s)}{n}) + O(\frac{1}{n^2}) + O(\frac{1}{n^{\alpha k}}))^m$$

$$= \binom{n}{ns}(n^\alpha - 1)^{n-S}(1-p)^{m \ln n}[1 + \frac{p}{1-p}(s^k + \frac{g(s)}{n})]^{m \ln n}(1 + O(\frac{1}{n}))$$

$$= n^{\alpha n}(1-p)^{rn \ln n}(1 - \frac{1}{n^\alpha})^{n-ns}(\frac{1}{n^\alpha})^{ns}\binom{n}{ns}[1 + \frac{p}{1-p}(s^k + \frac{g(s)}{n})]^{m \ln n}(1 + O(\frac{1}{n}))$$

$$= E(X_{r,p}^{n,k,\alpha})(1 - \frac{1}{n^\alpha})^{n-ns}(\frac{1}{n^\alpha})^{ns}\binom{n}{ns}[1 + \frac{p}{1-p}(s^k + \frac{g(s)}{n})]^{m \ln n}(1 + O(\frac{1}{n}))$$

When n is sufficiently large, *F(s)* is mainly determined by the following terms *f(s)*, excluding the first term $E(X_{r,p}^{n,k,\alpha})$:

$$f(s) = [1 + ps^k/(1-p)]^{rn \ln n}(1/n^\alpha)^{ns}$$

$$= e^{[r \ln(1 + ps^k/(1-p)) - \alpha s]n \ln n}$$

Let *h(s)* denote the index of the above formula, we have

$$h(s) = [r \ln(1 + ps^k/(1-p)) - \alpha s]n \ln n$$

The first and second derivatives of *h(s)* are:

$$h(s)' = \frac{rpks^{k-1}}{(1-p+ps^k)^2}$$

$$h(s)'' = \frac{rkps^{k-2}[(k-1)(1-p)-ps^k]}{(1-p+ps^k)^2}$$

Applying the condition $k \geq 1/(1-p)$ we can easily prove that $h(s'') \geq 0$ on the interval $0 \leq s \leq 1$. It is not hard to show that $h(0)=0$, and $h(1) < 0$ when $p < p_{cr}$. Hence we can easily prove that the unique maximum point of $h(s)$ is $s = 0$ when $p<p_{cr}$, and therefore the terms of $0 < s \leq 1$ are negligible. Now we only need to consider those terms near s=0. The process can be divided into the following cases:

(a1) $1/k < \alpha < 1$. Let $S_0 = n^\beta$ (where $\beta$ is a constant and satisfies $1-\alpha < \beta < 1-1/k$). It is not hard to show that when $0 \leq S \leq S_0$ ( $0 \leq s \leq n^{\beta-1} < n^{-1/k}$ ), the following limit holds:

$$\lim_{n \to \infty} p(s^k + \frac{g(s)}{n}) n \ln n / (1-p) \approx e^0 = 1$$

So when $0 \leq S \leq S_0$, the asymptotic estimate of $F(s)$ is

$$F(s) \approx E(X_{r,p}^{n,k,\alpha})\binom{n}{S}(1-1/n^\alpha)^{n-S}(1/n^\alpha)^S$$

It should be noted that $\binom{n}{S}(1-1/n^\alpha)^{n-S}(1/n^\alpha)^S$ is a binomial term whose maximum point is around $S = n^{1-\alpha}$, and $S_0 = n^\beta > n^{1-\alpha}$. By asymptotic analysis, we obtain

$$\sum_{S=0}^{S_0} \binom{n}{S}(1-\frac{1}{n^\alpha})^{n-S}(\frac{1}{n^\alpha})^S \approx \sum_{S=0}^{n} \binom{n}{S}(1-\frac{1}{n^\alpha})^{n-S}(\frac{1}{n^\alpha})^S = 1$$

Thus we get

$$E(X_{r,p}^{n,k,\alpha}|a=1) \approx E(X_{r,p}^{n,k,\alpha})$$

(b1) $\alpha > 1$. When $S = 0$, we have

$$\left[1+p/(1-p)(s^k+\frac{g(s)}{n})\right]^{m\ln n} = 1$$

then we get

$$F(s) \approx E(X_{r,p}^{n,k,\alpha})$$

Similarly, we have

$$F(s) \approx E(X_{r,p}^{n,k,\alpha})n^{1-\alpha} \quad \text{when } S = 1$$

$$F(s) \approx E(X_{r,p}^{n,k,\alpha})n^{2(1-\alpha)}/2! \quad \text{when } S = 2$$

…

Summing the above terms together, we obtain

$$E(X_{r,p}^{n,k,\alpha}|a=1) \approx E(X_{r,p}^{n,k,\alpha})$$

(c1) $\alpha = 1$. The analysis is similar as (b1), we have

$$F(s) \approx E(X_{r,p}^{n,k,\alpha})(0!e)^{-1} \quad \text{when } S = 0$$

$$F(s) \approx E(X_{r,p}^{n,k,\alpha})(1!e)^{-1} \quad \text{when } S = 1$$

$$F(s) \approx E(X_{r,p}^{n,k,\alpha})(2!e)^{-1} \quad \text{when } S = 2$$

…

Summing the above terms together, we obtain

$$E(X_{r,p}^{n,k,\alpha}|a=1) \approx E(X_{r,p}^{n,k,\alpha})$$

From above three cases (a1), (b1) and (c1), we have:

$$E(X_{r,p}^{n,k,\alpha}|a=1) \approx E(X_{r,p}^{n,k,\alpha}) \quad \text{when } p < p_{cr.} \tag{11}$$

Hence:

$$\lim_{n \to \infty} \frac{E(X_{r,p}^{n,k,\alpha})}{E(X_{r,p}^{n,k,\alpha}|a=1)} = 1, \text{ when } p < p_{cr}.$$

Since, $\Pr[X_{r,p}^{n,k,\alpha} \geq d^{\alpha/t}] \geq E(X_{r,p}^{n,k,\alpha}) / E(X_{r,p}^{n,k,\alpha} | a=1)$,

We have

$$\lim_{n \to \infty} \Pr[X_{r,p}^{n,k,\alpha} \geq d^{n/t}] = 1, \text{ when } p < p_{cr}$$

Hence, equation (1) stands, and the theorem is proved. □

According to theorem 1, when $p < p_{cr}$, the problem is under constrained, the solutions of the instances in this region are more than $d^{n/t}$; when $p > p_{cr}$, the problem is over constrained, the solutions of the instances are less than $d^{n/t}$; and the exact point is $p_{cr}$. Theorem 1 uses constraint tightness as the parameter, when use constraint density as the parameter, we get a similar result.

**Theorem 2** If $r_{cr} = -\dfrac{\alpha(1-\frac{1}{t})}{\ln(1-p)}, \alpha > \dfrac{1}{k}$ and $0<p<1$ are two constraints satisfy $k \geq 1/(1-p)$, then

$\lim_{n\to\infty} \Pr[X_{r,p}^{n,k,\alpha} \geq d^{n/t}] = 1$ when r<$r_{cr}$

$\lim_{n\to\infty} \Pr[X_{r,p}^{n,k,\alpha} \geq d^{n/t}] = 0$ when r>$r_{cr}$

The proof of Theorem 2 is similar to Theorem 1.

The above two theorems hold when n→∞, so we make some preliminary experiments to show that exact phase transition do occur even if the number of variables is small. The experiments are conducted on a machine with (2.4 GHZ CPU and 2G memory). We encode CSP instances into SAT instances, and use *Cachet* (Sang et. al. 2004) to obtain exact solutions numbers. It usually takes Cachet hours to solve a CSP instance with more than 15 variables, so we only conduct experiments on small CSP instance. In Fig. 1, the parameters of RB model are set as: k=2, α=0.8, r=1.7, and *n*={7, 10, 13}. And in Fig. 2 the parameters are set to be α=0.85, *r*=1.4, n={9, 12, 15}. For each point we generate 100 instances. In both figures, we consider the problem of #CSP($\geq d^{n/2}$). So in Fig. 1, the threshold numbers of solutions are 280, 7776, and 741460 respectively, and in Fig. 2, the threshold numbers of solutions are 3175, 262144, and 3.1623e+007 respectively. According to theorem 1, in Fig. 1, $p_{cr} \approx 0.210$, and in Fig. 2, $p_{cr} \approx 0.262$. Note that even if n is relatively small, the phase transition phenomenon is really obvious, and the practical threshold is near to the theoretical threshold. We also conduct experiments to validate theorem 2 and get similar results. We do not observe obvious phase transitions of mean CPU time in our experiments. The reason may be that Cachet is a counting solver rather than a #SAT($\geq d^{n/2}$) solver. We guess that phase transitions for CPU time may exist for some modified version of exact counting solvers.

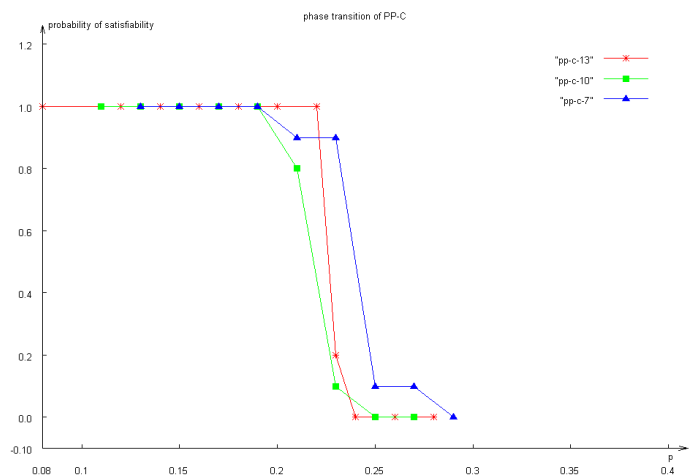

Figure 1: Phase transition for (2,{7,10,13},0.8,1.7,*p*)

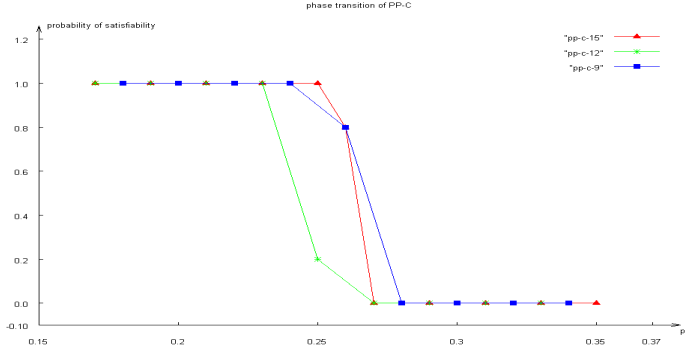

Figure 2: Phase transition for (2,{9,12,15},0.85,1.4,*p*)

## On Estimating Solutions Number of CSPs

In the former section, we have proved that exact phase transitions for #CSP($\geq d^{n/t}$) occur for constraint satisfaction problems in Model RB. When t goes to infinity, the problem reduces to #CSP($\geq 1$), i.e. deciding satisfiability of CSP, and the theorems proposed in this paper also reduce to theorem 1 and theorem 2 in (Xu and Li 2000). In this sense, the theorems proposed in this paper can be viewed as a generalization of those proposed in (Xu and Li 2000).

In this section, we further show that a careful analysis of the phase transition can lead to an approximate algorithm to estimate the solutions number in Model RB while applying a limited restriction on domain size and on constraint tightness. First, when t and n goes to infinity, according to theorem 1 (or theorem 2), we know that if $p > p_{cr} = 1 - e^{-\alpha/r}$ (or $r > r_{cr} = -\alpha/\ln(1-p)$), the number of solutions of *I* approaches 0. So we only need to consider conditions when $p < p_{cr}$ or $r < r_{cr}$. Recall in formula (1), we have shown that the expected number of solutions E($X_{r,p}^{n,k,\alpha}$) of the instance can be calculated as:

$$E(X_{r,p}^{n,k,\alpha}) = d^n(1-p)^{rn\ln n} = n^{\alpha n}(1-p)^{rn\ln n}.$$

Now we focus on proving that E($X_{r,p}^{n,k,\alpha}$) can act as a good estimation for Model RB.

**Theorem 3** Given a CSP instance *I* randomly generated following Model RB and *k*, $\alpha$ *and r* satisfy the inequality $k \cdot e^{-\alpha/r} \geq 1$ ($k \geq 1/(1-p), resp.$) and $\alpha > 1/k$. Let $X_{r,p}^{n,k,\alpha}$ be the number of solutions for *I*, E($X_{r,p}^{n,k,\alpha}$) denote the expected number of solution for *I*, $\delta$ be an arbitrary real number, and *Pr* denote the function for probability distribution. As the number of variable goes to infinity, if $p < 1 - e^{-\alpha/r}$ ($r < -\alpha/\ln(1-p), resp.$), then

$$\underset{n\to\infty}{Lim}(Pr((1-\delta)E(X_{r,p}^{n,k,\alpha}) < X_{r,p}^{n,k,\alpha} < (1+\delta)E(X_{r,p}^{n,k,\alpha}))) \approx 1$$

*Proof.* According to the Chebyshev inequality, we have

$$P(|X_{r,p}^{n,k,\alpha} - E(X_{r,p}^{n,k,\alpha})| < \varepsilon) \geq 1 - \frac{D(X_{r,p}^{n,k,\alpha})}{\varepsilon^2} \qquad (12)$$

We can rewrite (12) to get

$$P(E(X_{r,p}^{n,k,\alpha})-\varepsilon < X_{r,p}^{n,k,\alpha} < E(X_{r,p}^{n,k,\alpha})+\varepsilon) \geq 1 - \frac{D(X_{r,p}^{n,k,\alpha})}{\varepsilon^2} \qquad (13)$$

According to the definition of $D(X_{r,p}^{n,k,\alpha})$, we have

$$D(X_{r,p}^{n,k,\alpha}) = E((X_{r,p}^{n,k,\alpha})^2) - E^2(X_{r,p}^{n,k,\alpha}) \qquad (14)$$

Define

$$\varepsilon = \delta \cdot E(X_{r,p}^{n,k,\alpha}) \qquad (15)$$

By combining formula (13), (14) and (15),

$$Pr(E(X_{r,p}^{n,k,\alpha})-\delta E(X_{r,p}^{n,k,\alpha}) < X_{r,p}^{n,k,\alpha} < E(X_{r,p}^{n,k,\alpha})+\delta E(X_{r,p}^{n,k,\alpha}))$$

$$\geq 1 - \frac{E((X_{r,p}^{n,k,\alpha})^2) - E^2(X_{r,p}^{n,k,\alpha})}{\delta^2 E^2(X_{r,p}^{n,k,\alpha})} \qquad (16)$$

By rewriting formula, we get

$$Pr((1-\delta)E(X_{r,p}^{n,k,\alpha}) < X_{r,p}^{n,k,\alpha} < (1+\delta)E(X_{r,p}^{n,k,\alpha}))$$

$$\geq 1 - \frac{E((X_{r,p}^{n,k,\alpha})^2) - E^2(X_{r,p}^{n,k,\alpha})}{\delta^2 E^2(X_{r,p}^{n,k,\alpha})} \qquad (17)$$

In theorem 1 of (Xu and Li 2000), it has been proved that $E((X_{r,p}^{n,k,\alpha})^2) \approx E^2(X_{r,p}^{n,k,\alpha})$ when n is large enough.

So, the theorem is proved, and we have

$$\underset{n\to\infty}{Lim}(Pr((1-\delta)E(X_{r,p}^{n,k,\alpha}) < X_{r,p}^{n,k,\alpha} < (1+\delta)E(X_{r,p}^{n,k,\alpha}))) \approx 1 \quad \square$$

From above discussion, we can easily develop an approximate algorithm to estimate the solution numbers in linear time, as we called AE-count. The only point where AE-count cannot work is when $p = p_{cr}$. Note that, until now, state-of-art exact counting algorithms such as *Relsat* (Bayardo and Pehoushek 2000) and *Cachet* (Sang et. al. 2004) cannot scale up for large scale problems. For approximate algorithms, although the computational time is relatively acceptable, few of them can provide good bounds on the solutions numbers with high confidence (Gomes et. al. 2006). Moreover, for large scale counting problems, even for the approximate algorithms, the solution processes are always time consuming. It should be pointed out that the short coming of AE-count is that it is only capable of estimating random instances generated following Model RB, and can only return the estimate numbers of solutions rather enumerate the solutions. However, since we can generate large-scale CSP instances or SAT instances following Model RB, and simultaneously obtain an upper bound and lower bound of their solution numbers, we can use model RB to generate benchmarks for counting algorithms.

To test the accuracy of AE-count, the estimated solutions number is compared with the exact solution numbers, as can be seen in Table 1. Here $\alpha$, $r$, $p$ and $n$ denote domain density, constraint density, constraint tightness and variables numbers respectively. The real number $\delta$ is from 0.5 to 0.9 in step of 0.1, so there are five intervals. Experimental results in (Xu et. al. 2007) have shown SAT-UNSAT phase transitions do occur. So we only generate random instances in the SAT region, and for each point 300 instances are generated. To get the exact solutions numbers, we encode CSP instances into SAT instances, and then use *Cachet* to count the models. For the computational limit of *Cachet*, we only test instances with no more than 20 variables. As can be seen in Table 1, even if the problems scales are relatively small, the estimates are found to be over 69% correct for $\delta = 0.9$, and the accuracy of the estimates grows with the increasing of problem scales. Table 2 shows the comparisons of solutions numbers between Cachet, Approx-count, and AE-count. For the solutions number in AE-count, we use $E(X_{r,p}^{n,k,\alpha})$ instead of an interval. Each number in the table is an average of the solutions numbers of 300 random instances in the SAT region.

| $\alpha$ | $r$ | $p$ | $n$ | [0.5,1.5] | [0.4,1.6] | [0.3,1.7] | [0.2,1.8] | [0.1,1.9] |
|---|---|---|---|---|---|---|---|---|
| 0.7 | 2.3 | 0.2 | 13 | 23.33% | 26.67% | 26.67% | 33.33% | 36.67% |
| 0.7 | 2.3 | 0.2 | 14 | 40.00% | 43.33% | 60.00% | 70.00% | 76.67% |
| 0.7 | 2.3 | 0.2 | 15 | 36.67% | 43.33% | 50.00% | 50.00% | 50.00% |
| 0.7 | 2.3 | 0.2 | 16 | 50.00% | 70.00% | 70.00% | 76.67% | 83.33% |
| 0.8 | 1.5 | 0.3 | 7 | 46.67% | 63.33% | 76.67% | 83.33% | 83.33% |
| 0.8 | 1.5 | 0.3 | 9 | 66.67% | 76.67% | 76.67% | 83.33% | 83.33% |
| 0.8 | 1.5 | 0.3 | 11 | 63.33% | 70.00% | 76.67% | 83.33% | 83.33% |
| 0.8 | 1.5 | 0.3 | 13 | 36.67% | 40.00% | 43.33% | 53.33% | 56.67% |
| 0.9 | 2.1 | 0.3 | 11 | 6.67% | 6.67% | 13.33% | 13.33% | 13.33% |
| 0.9 | 2.1 | 0.3 | 12 | 50.00% | 70.00% | 80.00% | 86.67% | 90.00% |
| 0.9 | 2.1 | 0.3 | 13 | 66.67% | 70.00% | 70.00% | 70.00% | 73.33% |
| 0.9 | 2.1 | 0.3 | 14 | 33.33% | 40.00% | 43.33% | 46.67% | 53.33% |
| 1 | 2 | 0.35 | 11 | 46.67% | 50.00% | 63.33% | 63.33% | 66.67% |
| 1 | 2 | 0.35 | 13 | 53.33% | 60.00% | 80.00% | 83.33% | 86.67% |
| 1 | 2 | 0.35 | 15 | 66.67% | 66.67% | 70.00% | 83.33% | 86.67% |
| 1 | 2 | 0.35 | 17 | 66.67% | 70.00% | 73.33% | 80.00% | 83.33% |

Table 1 Accuracy of AE-count

| $\alpha$ | 0.6 | 0.7 | 0.8 | 1 |
|---|---|---|---|---|
| $r$ | 1.8 | 2.3 | 1.5 | 2 |
| $n$ | 17 | 16 | 9 | 17 |
| $p$ | 0.21 | 0.2 | 0.3 | 0.31 |
| Cachet | 5091.1 | 4443.1 | 253.6 | 247212.5 |
| Approx-count | 10338 | 5547.7 | 487.8 | 6428.1 |
| AE-count | 4734 | 39655 | 188.7 | 247790 |

Table 2 Comparisons of solutions numbers

# Conclusion

In this paper, we first present a probabilistic analysis of the problem #CSP. We show that for random instances generated following Model RB, exact phase transitions do exist for a decision version of #CSP, i.e. #CSP($\geq d^{n/t}$). Then, preliminary experimental results have confirmed the phase transition and threshold predicted by theory. Second, we show through a careful analysis of phase transition, we can present an accurate estimate of solution numbers of random CSP instances generated following Model RB. Such results should be valuable for future counting algorithms. First, we can generate large scale instances with upper and lower bounds of solution numbers. Second, our approach may be an inspiration to develop similar approximate counting algorithms for other random models.